\documentclass[journal=jacsat,manuscript=article]{achemso}

\usepackage{chemformula} 
\usepackage[T1]{fontenc} 
\usepackage{adjustbox}
\usepackage[font=footnotesize,labelfont=bf,labelsep=period,width=0.75\textwidth]{caption}   
\usepackage[font=footnotesize,labelfont=bf,labelsep=period]{subcaption}                     
\usepackage{lscape}

\DeclareCaptionSubType*[arabic]{figure}                                                     
\DeclareCaptionLabelFormat{subfiglabel}{Figure #2}                                          
\usepackage{amsmath}
\usepackage{bm}

\usepackage{booktabs}
\usepackage{adjustbox}

\title{ Multimodal machine learning with large language embedding model for polymer property prediction }

\author{Tianren Zhang}
\affiliation[University of Delaware]
{Department of Materials Science and Engineering, University of Delaware, Newark, Delaware 19716, United States}
\alsoaffiliation[University of Pennsylvania]
{Department of Chemistry, University of Pennsylvania, Philadelphia, Pennsylvania 19104, United States}
\email{tianren@udel.edu}

\author{Dai-Bei Yang}
\affiliation[University of Pennsylvania]
{Department of Chemistry, University of Pennsylvania, Philadelphia, Pennsylvania 19104, United States}

\begin{document}
\newpage

\begin{abstract}
Contemporary large language models (LLMs), such as GPT-4 and Llama, have harnessed extensive computational power and diverse text corpora to achieve remarkable proficiency in interpreting and generating domain-specific content, including materials science. To leverage the domain knowledge embedded within these models, we propose a simple yet effective multimodal architecture, PolyLLMem. By integrating text embeddings from Llama 3 with molecular structure embeddings from Uni-Mol, PolyLLMem enables the accurate prediction of polymer properties. Low-Rank Adaptation (LoRA) layers were integrated into our model during the property prediction stage to adapt the pretrained embeddings to our limited polymer dataset, thereby enhancing their chemical relevance for polymer SMILES representation. Such a balanced fusion of fine-tuned textual and structural information enables PolyLLMem to robustly predict a variety of polymer properties despite the scarcity of training data. Its performance is comparable to, and in some cases exceeds, that of graph-based or transformer-based models that typically require pretraining on millions of polymer samples. These findings demonstrate that LLM, such as Llama, can effectively capture chemical information encoded in polymer PSMILES, and underscore the efficacy of multimodal fusion of LLM embeddings and molecular structure embeddings in overcoming data scarcity and accelerating the discovery of advanced polymeric materials.

\end{abstract}
\newpage

\section{Introduction}

Polymeric materials with their complex architectures and diverse functionalities serve as good candidates for a wide array of applications ranging from everyday consumer products to advanced lightweight aerospace and biomedical devices. \cite{ornaghi2023review,oladele2020polymer} Their unique properties, such as high molecular weight, tunable chemical functionality, and versatile mechanical behavior, enable the design of materials that are tailored to specific performance requirements. Accurate prediction of polymer properties can significantly accelerate the materials discovery process, allowing researchers to rapidly identify and optimize promising candidates while reducing reliance on time-consuming and costly experimental trials.\cite{doan2020machine,mcdonald2023applied,sharma2022advances,xu2022new,ge2025machine,patra2021data}

Despite being intriguing, the intrinsic complexity of polymer structures, coupled with the limited size of available databases, poses significant challenges for accurate property prediction. To overcome these limitations, researchers in the polymer field have adopted a variety of strategies, including advancements in polymer representation, feature extraction, and data augmentation, to enhance the performance of diverse machine learning (ML) architectures and accelerate the development of predictive models in polymer research. For instance, similar to Simplified Molecular-Input Line-Entry System (SMILES)\cite{Weininger1988} used for small molecules, polymer SMILES including Polymer Simplified Molecular-Input Line-Entry System (PSMILES) and BigSMILES\cite{lin2019bigsmiles,kuenneth2023polybert} were proposed as an extension of the traditional SMILES notation to describe macromolecular structures, including repeating units, end groups, and connectivity patterns. Furthermore, various molecular descriptors and structural representations, such as Morgan fingerprint and its frequency-based variant, as well as molecular graphs \cite{kim2018polymer,Tao2021,wu2019machine,qiu2024polync,uddin2024interpretable,zhao2023review}, have been derived from polymers and utilized as input features for ML models. Those notations and feature extractions enable the standardized digital representation of polymers, facilitating computational analysis, database storage, and interoperability in polymer informatics applications. Additionally, to overcome the limitations imposed by the small size of polymer databases, new datasets were generated. Examples included the PI1M database, constructed first by training a generative model on approximately 12,000 polymers manually collected from PolyInfo, subsequently generating around one million hypothetical polymers \cite{ma2020pi1m}. Similarly, a dataset of 100 million hypothetical polymers was created by enumeratively combining chemical fragments extracted from over 13,000 synthesized ones.\cite{kuenneth2023polybert} 

With these advances in data preparation, both classical ML and deep-learning models have been employed to enhance polymer property prediction. Classical models, such as ensemble tree-based methods, support vector regression, and Gaussian processes, have achieved fair performance in predicting properties on glass transition temperature ($T_g$) datasets when paired with chemically informed descriptors like Morgan fingerprints and RDKit features \cite{doan2020machine,volgin2022machine,landrum2013rdkit,rogers2010extended,uddin2024interpretable,wu2019machine}.On the deep learning side, architectures including Graph Neural Networks (GNNs) and transformer-based models have been used to learn directly from polymer structures or PSMILES representations\cite{gilmer2017neural,wilson2023polyid,yue2023high,tao2023discovery,zeng2018graph,huang2024enhancing,gurnani2023polymer}. Recent models such as ChemBERTa, TransPolymer, and polyBERT leverage large-scale pretraining on polymer SMILES to generate contextual embeddings for downstream property prediction tasks\cite{xu2023transpolymer,kuenneth2023polybert,zhang2023transferring,chithrananda2020chemberta}. Other recent frameworks further enhance performance through multimodal fusion, integrating structural, textual, or generative information\cite{qiu2024polync,wang2024mmpolymer}. Although these approaches demonstrated high accuracy in the predictions of polymer properties, they typically required careful architecture design and large volumes of real or virtual polymer data for pretraining before being effectively applied to downstream tasks.

One of the most remarkable developments in machine learning is the emergence of large language models (LLMs), which have leveraged vast computational resources and extensive text corpora, including materials science literature, to demonstrate exceptional capabilities in understanding, reasoning, and generating domain-specific content.\cite{achiam2023gpt,grattafiori2024llama,touvron2023llama,jia2024llmatdesign,luu2024bioinspiredllm,stewart2024molecular}. Although originally developed for general-purpose natural language tasks, LLMs have shown strong potential in scientific domains when appropriately prompted or minimally tuned. For example, GenePT leveraged LLM-derived embeddings from gene descriptions and single-cell data, achieving performance comparable to or exceeding models pretrained on gene-expression profiles from millions of cells for tasks like gene-property and cell-type classification\cite{chen2024simple}. Similarly, LLMs, such as Llama, could also have been exposed to substantial polymer-related knowledge during pretraining and therefore, be capable of extracting relevant information from polymer texts for downstream tasks such as property prediction. Moreover, to complement the textual embeddings and further enhance predictive performance, we incorporated structural embeddings from Uni-Mol, a deep-learning model designed to encode detailed molecular structures\cite{zhou2023uni}. Pretrained on millions of 3D molecular representations of small molecules, Uni-Mol could effectively capture essential chemical and spatial features relevant to prediction tasks. 

By integrating embeddings from both the LLM and Uni-Mol, we developed PolyLLMem, a multimodal neural network tailored specifically for polymer property prediction. In our approach, the complementary information from both textual and structural domains was balanced, and a Low-rank adaptation (LoRA) layer was incorporated to fine-tune the embeddings with our target small polymer dataset during the property prediction tasks. We evaluated PolyLLMem on predictions of 22 polymer properties, and our results revealed that the integrated approach yields performance comparable to, and in some cases exceeds, that of graph-based or transformer-based models that typically require pretraining on millions of polymer samples. The PolyLLMem offers several advantages: (i) it demonstrates robust performance across a variety of property prediction tasks, even when trained on limited data; (ii) it requires minimal dataset curation, preprocessing, or additional pretraining on polymer-specific corpora; and (iii) it is computationally efficient and straightforward to implement, making it accessible for broad application.

\newpage
\section{Method}

\subsubsection{Data Collection} Our dataset comprises 29,639 data points of homopolymers covering 22 properties obtained from both DFT calculations and experimental measurements, sourced from peer-reviewed literatures and established databases \cite{huan2015accelerated,huan2016polymer,sharma2014rational,gurnani2023polymer,otsuka2011polyinfo,afzal2020high,kuenneth2021polymer,phan2024gas}. The properties, including glass transition temperature ($T_g$), melting temperature ($T_m$),  thermal decomposition temperature ($T_d$), atomization energy ($E_{at}$) , crystallization tendency ($X_c$), density ($\rho$), band gap (chain) ($E_{gc}$), band gap (bulk) ($E_{gb}$), electron affinity ($E_{ea}$), ionization energy ($E_i$), refractive index ($n_c$), conductivity ($\sigma$), tensile strength at yield ($\sigma_y$), Young's modulus ($E$), tensile strength at break ($\sigma_b$), elongation at break ($\epsilon_b$), and gas permeability of O$_2$, CO$_2$, N$_2$, H$_2$, He, CH$_4$  ($\mu_{O_2}$, $\mu_{CO_2}$, $\mu_{N_2}$, $\mu_{H_2}$, $\mu_{He}$ and $\mu_{CH_4}$). The detailed distribution range for each property is provided in Table~\ref{tab:polymer_data} and Supplementary Information (SI). Due to the extensive range observed in gas permeability values, mechanical-related properties, and conductivity, which span several orders of magnitude, a base-10 logarithmic transformation was applied to $\mu_{O_2}$, $\mu_{CO_2}$, $\mu_{N_2}$, $\mu_{H_2}$, $\mu_{He}$, $\mu_{CH_4}$, $\sigma_y$, $E$, $\sigma_b$, $\epsilon_b$ and $\sigma$ to normalize their distributions and stabilize the variance. Finally, the polymer dataset was split into training and testing sets using an 85/15 ratio, with the testing set reserved solely for final property prediction.

\begin{table}[htbp]
    \centering
    \caption{Polymer property dataset with sources, data ranges, and data points. The data set contains 22 properties for homo polymers.}
    \label{tab:polymer_data}
    \resizebox{\textwidth}{!}{%
    \begin{tabular}{lllllr}
    \hline
    \textbf{Property} & \textbf{Symbol} & \textbf{Unit} & \textbf{Source} & \textbf{Data Range} & \textbf{Data Points} \\
    \hline
    Glass transition temp. & $T_g$ & \textdegree C & Exp. & [-1.2e+02, 5e+02] & 6769 \\
    Melting temp. & $T_m$ & \textdegree C & Exp. & [-5.5e+01, 5.8e+02] & 3349 \\
    Thermal decomposition temp. & $T_d$ & \textdegree C & Exp. & [1.8e+01, 8.5e+02] & 5347 \\
    Atomization energy & $E_{at}$ & eV atom$^{-1}$ & DFT & [-7e+00, -5e+00] & 390 \\
    Crystallization tendency (DFT) & $X_c$ & \% & DFT & [1e-01, 1e+02] & 432 \\
    Density & $\rho$ & g cm$^{-3}$ & Exp. & [1e-01, 3e+00] & 1520 \\
    Band gap (chain) & $E_{gc}$ & eV & DFT & [2e-02, 1e+01] & 3380 \\
    Band gap (bulk) & $E_{gb}$ & eV & DFT & [4e-01, 1e+01] & 561 \\
    Electron affinity & $E_{ea}$ & eV & DFT & [4e-01, 5e+00] & 368 \\
    Ionization energy & $E_i$ & eV & DFT & [3.5e+00, 1e+01] & 370 \\
    Refractive index & $n_c$ & - & DFT & [1e+00, 3e+00] & 382 \\
    Conductivity & $\sigma$ & S/cm & Exp. & [0e+00, 1e+07] & 382 \\
    Young's modulus & $E$ & GPa & Exp. & [2e-05, 6e+00] & 938 \\
    Tensile strength at yield & $\sigma_y$ & GPa & Exp. & [3e-08, 4e-01] & 244 \\
    Tensile strength at break & $\sigma_b$ & GPa & Exp. & [8e-05, 2e-01] & 975 \\
    Elongation at break & $\epsilon_b$ & - & Exp. & [6e-01, 1e+03] & 1015 \\
    O$_2$ gas permeability & $\mu_{O_2}$ & barrer & Exp. & [3e-04, 1.9e+04] & 695 \\
    CO$_2$ gas permeability & $\mu_{CO_2}$ & barrer & Exp. & [1e-03, 4.7e+04] & 644 \\
    N$_2$ gas permeability & $\mu_{N_2}$ & barrer & Exp. & [1e-04, 1.7e+04] & 678 \\
    H$_2$ gas permeability & $\mu_{H_2}$ & barrer & Exp. & [2e-02, 3.7e+04] & 461 \\
    He gas permeability & $\mu_{He}$ & barrer & Exp. & [5e-02, 1.8e+04] & 408 \\
    CH$_4$ gas permeability & $\mu_{CH_4}$ & barrer & Exp. & [4e-04, 3.5e+04] & 331 \\
    \hline
    \end{tabular}
    }
    \label{tab:polymer_data}
\end{table}

\subsubsection{Model Architecture} 
\begin{figure}[h]
\centering
\includegraphics[width=1.0\textwidth]{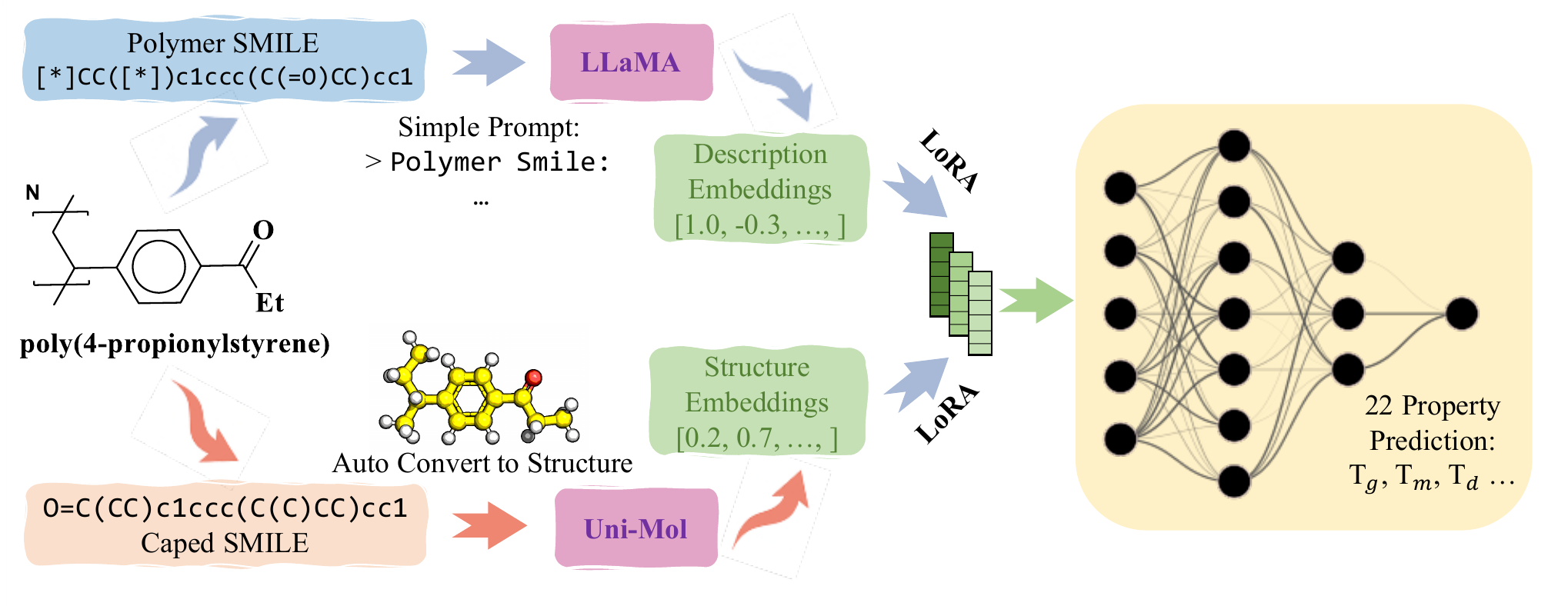}
\caption{Schematic representation of PolyLLMem architecture. The architecture processes two inputs: text-based PSMILES string is encoded using the Llama3 model to generate embeddings, while 3D molecular representations are processed automatically by Uni-Mol to extract structural embeddings from a caped SMILES. These embeddings are merged after LoRA layers to form a unified representation that is subsequentially employed in a single-task framework with a Multilayer Perceptron (MLP) for training and predicting polymer properties.}
\label{fig:arch}
\end{figure}

Our multimodal model (PolyLLMem) integrated LLM-based and Uni-Mol-based embeddings to predict polymer properties by capturing both textual and structural information. The LLM-based embeddings were generated using the LLM Llama3 by mean pooling the final hidden states (token-level embeddings) to produce a single embedding vector with 4096 dimensions for each input textual string\cite{grattafiori2024llama}. Each input string, as shown in Figure.~\ref{fig:arch} follows the format “Polymer Smile: PSMILES.” (Figure\ref{fig:arch}) The mean pooling of the token-level embeddings effectively distills the rich chemical context captured by the LLM-based model into a robust representation.

In parallel, we employed Uni-Mol embeddings, which were 1536-dimensional, to capture the 3D geometry and conformational details of the molecules. This approach yielded embeddings that encapsulate critical geometric relationships, providing complementary structural insights to the text-based representations obtained from the Llama3. Moreover, since Uni-Mol does not recognize PSMILES, we replaced the asterisk \texttt{*} with "\texttt{C}" in the input for Uni-Mol (caped SMILES, Figure \ref{fig:arch} ). Once both embeddings were obtained, each was projected into a common latent space with a predefined hidden size using a linear layer with Gaussian Error Linear Unit (GELU) activation and batch normalization. LoRA layers further refine these projections, and a gated fusion mechanism dynamically combines the updated LLM and Uni-Mol representations.\cite{hu2022lora}. The fused embeddings were subsequently processed through a refinement block and a dedicated regression network, where each network was responsible for predicting a single target property.

\subsubsection{Training} Our training procedure employed 5-fold cross-validation to ensure a robust evaluation of the model’s predictive performance. Training was performed using a weight-decay–regularized optimizer alongside a learning rate schedule that adaptively reduces the step size when validation performance plateaus. An early stopping mechanism was applied to prevent overfitting.\cite{loshchilov2017decoupled,pedregosa2011scikit} Multiple loss functions, including Mean Square Error (MSE), Mean Absolute Error (MAE), and Huber, were used to guide optimization. For each fold, the best-performing model checkpoint was saved based on the validation loss, and final performance metrics (MAE and $R^2$) were computed on the test set and averaged across folds. Additionally, a grid search was used for hyperparameter tuning, optimizing key parameters such as hidden size, batch size, dropout rate, rank, alpha, learning rate, and weight decay.

For baseline comparisons, we evaluated a suite of classical ML models, including Random Forest (RF), Linear Regression (LR), Support Vector Regression (SVR), Decision Tree (DT), Ridge Regression (RR), AdaBoost, XGBoost, and a multilayer perceptron (MLP). Comparison was done using two distinct sets of input features separately: molecular descriptors (200 computed molecular properties) and Morgan fingerprints (MF) obtained from the RDKit package.\cite{landrum2013rdkit}. Additionally, embeddings generated by the Llama3 and Uni-Mol were separately evaluated using classical ML models to understand their contributions.

\newpage
\section{Discussions}
\subsubsection{Rich Information in Embeddings}
\begin{figure}[h]
\centering
\includegraphics[width=0.9\textwidth]{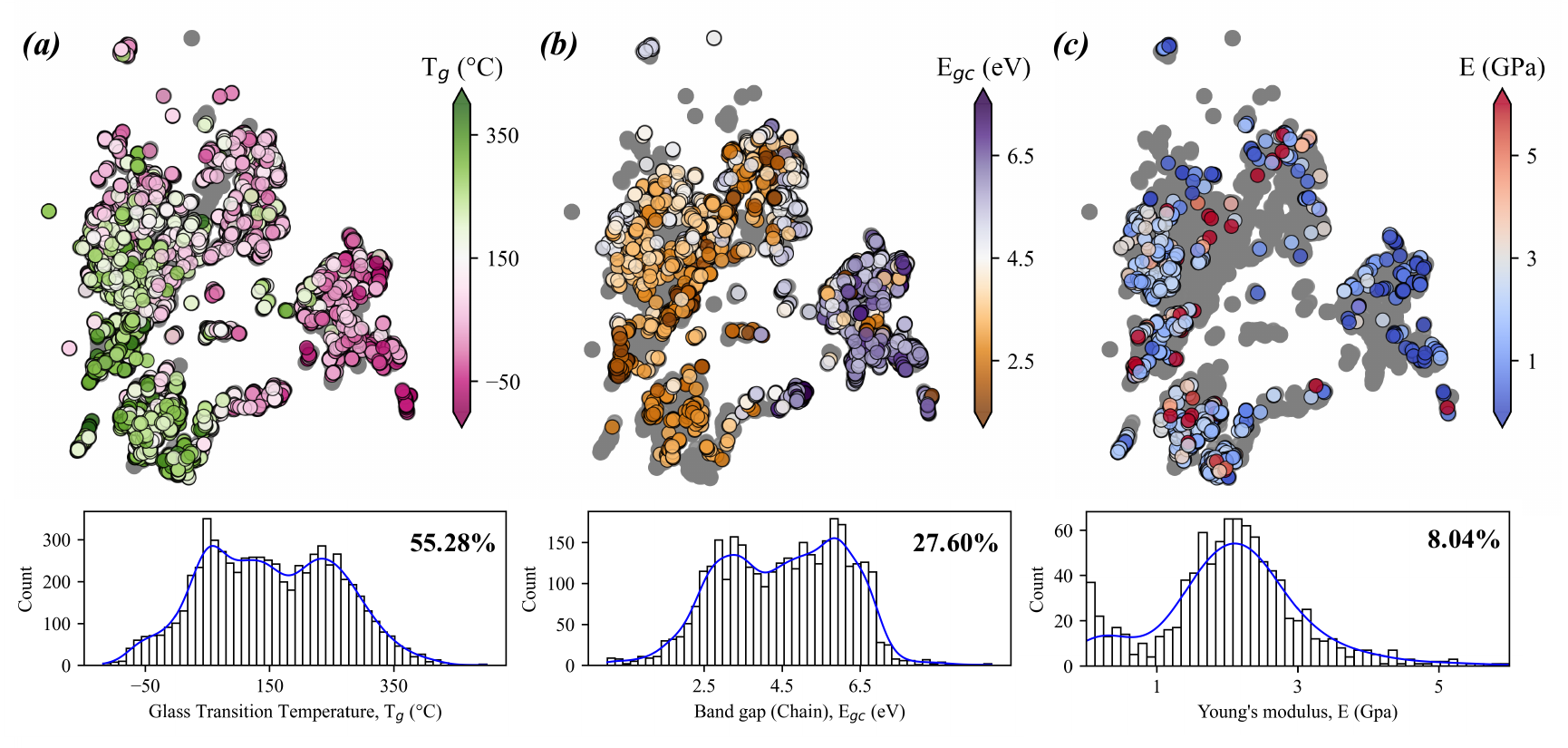}
\caption{ Upper: Two-dimensional Uniform Manifold Approximation and Projection (UMAP) plots of Llama3-generated polymer embeddings. The embeddings were first reduced to 100 dimensions using principal component analysis (PCA), followed by UMAP to project them into two dimensions. Panels (a–c) display colored dots representing property values for glass transition temperature ($T_g$), chain band gap ($E_{gc}$), and Young's modulus ($E$), respectively. Colors represent the value of the property, while light gray dots indicate polymers with missing values. In both (a) \& (b), A clear clustering of similar colors can be observed in each case, indicating that the LLM embeddings already capture meaningful chemical distinctions related to these properties prior to any task-specific training. Lower: Distributions of available data for each property. The proportion of known values relative to the entire dataset is also indicated.  }
\label{fig_umap}
\end{figure}

To evaluate the feasibility of using the LLM embedding model for polymer property predictions, we employed two-dimensional Uniform Manifold Approximation and Projection (UMAP) \cite{mcinnes2018umap} to visualize the generated embeddings for all polymers in this study (see Figure~\ref{fig_umap}).
These embeddings were obtained by mean pooling the final layer of the Llama3. In the UMAP plots, colored dots represent polymers with known property values for $T_{g}$ (Figure~\ref{fig_umap}a), $E_{gc}$ (Figure~\ref{fig_umap}b), and $E$ (Figure~\ref{fig_umap}c), while light gray dots indicate polymers with unknown property values. In each plot, polymers with similar property values tend to form localized clusters of similar colors, with the exception of the Young's modulus (Figure~\ref{fig_umap}c), where the differentiation is less distinct compared to the other properties. Nonetheless, this observation is noteworthy as it suggests that the Llama3 has successfully retained key chemical information and relationships inherent in the PSMILES strings, even though its predictive performance for properties like Young's modulus (Figure~\ref{fig_umap}c) may not be as robust as for others, potentially due to the low availability of these data.
\begin{figure}[H]
\centering
\includegraphics[width=1.0\textwidth]{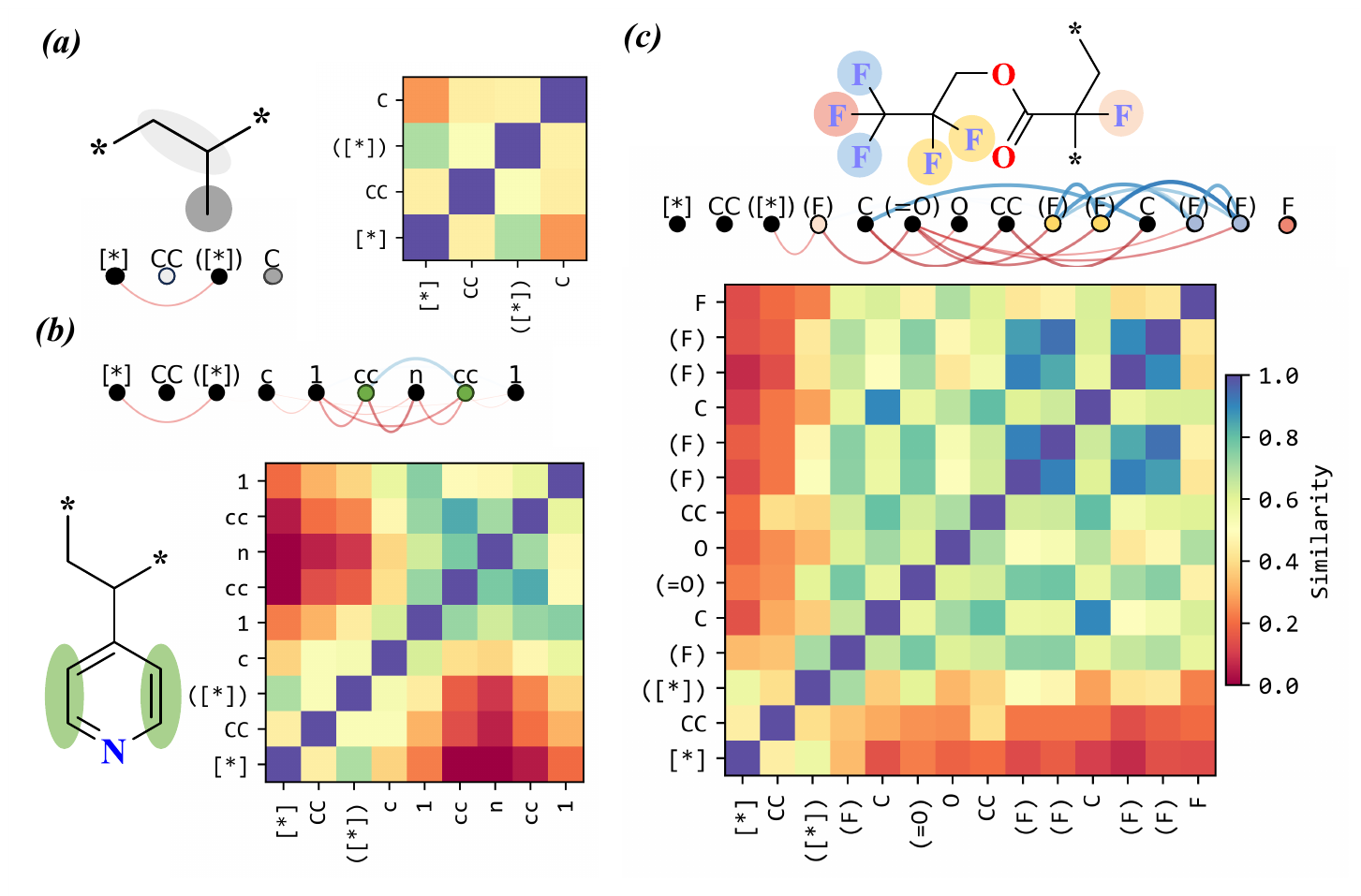}
\caption{Cosine similarity was computed between token-level embeddings for three representative polymers, with values close to 1 indicating high embedding similarity and potential shared chemical or structural features. Selected polymers (a--c): \texttt{[*]CC([*])C}, \texttt{[*]CC([*])\allowbreak c1ccncc1}, \texttt{[*]CC([*])(F)C(=O)\allowbreak OCC(F)(F)C(F)(F)F}.  Each example includes the molecular structure, a heatmap showing the full pairwise similarity matrix, and a chord diagram emphasizing strong inter-token relationships. In the chord diagrams, edges are drawn for token pairs exceeding a similarity threshold of 0.5 (0.7 used in c, for clarity). Different colors of nodes are used to show token locations. Line width represents similarity strength: blue edges connect tokens of the same character, while red edges indicate tokens of various names.  }
\label{fig:token_similarity}
\end{figure}

UMAP projections revealed that embeddings from the Llama3 encoded basic domain knowledge related to certain properties. However, to ensure this knowledge is generalizable, the embeddings must also capture underlying chemical features—such as symmetry, similarity, and structural relationships—beyond property-specific patterns. Here, we extracted token-level embeddings to verify such ability. Rather than using the embeddings after mean pooling for UMAP visualization, we retained the individual token representations prior to mean pooling to compute the cosine similarity across all tokens for the selected polymer. The approach began by tokenizing the PSMILES using a Llama3 tokenizer, originally built for natural language, which occasionally divides chemical notations into segments that do not inherently represent meaningful chemical substructures. For example, a chemical group like “\texttt{[*]}” might be split into separate tokens (e.g., “\texttt{[}” and “\texttt{*]}”), leading to dispersed embeddings that hinder straightforward interpretability. To address this, after obtaining the initial embedding output for each token achieved from Llama3 tokenizer, we implemented a custom embedding merging strategy where token representations that should collectively denote a single structural unit were averaged together. This involved selectively combining specific token embeddings, such as merging adjacent tokens or even portions of tokens, to realign the representation with the underlying polymer structure. Then the refined token level embeddings were used to calculate the cosine similarity among the refined tokens as shown in Figure~\ref{fig:token_similarity}. In all the panels in Figure~\ref{fig:token_similarity}a-c, tokens used for polymer notation\texttt{[*]}, tend to have moderate similarity scores (0.7) with each other, reflecting the model’s understanding of how they are used in PSMILES to denote branching, repeating units, or unspecified substituents. In contrast, lower similarity scores appear between tokens that represent clearly different chemical entities or notational functions (e.g., tokens related to side chains vs.\texttt{[*]} tokens). This separation indicates that the embeddings capture meaningful distinctions in chemical context.

In Figure~\ref{fig:token_similarity}a, the backbone token \texttt{CC} and the sidechain token \texttt{C} have significantly lower similarities, despite both being carbons. In Figure~\ref{fig:token_similarity}b, the PSMILES fragment consists of tokens such as \texttt{c}, \texttt{1}, \texttt{cc}, \texttt{n}, \texttt{cc}, and \texttt{1}, which collectively form a pyridine ring (e.g., \texttt{c1ccncc1}). These tokens show moderate inter-token similarity, indicating that the model recognizes them as parts of a unified aromatic ring structure and effectively captures their ring connectivity.  The symmetry of molecular structure is also preserved, exemplified by the tokens of cc. In Figure~\ref{fig:token_similarity}c, the presence of clear clustering for fluorine tokens, along with the separation from purely carbon-based tokens, demonstrates the Llama3’s ability to encode chemically relevant distinctions. Fluorine substituents have well-known effects on polymer properties (e.g., polarity, thermal stability), and the LLM embeddings’ separation of \texttt{F} from other tokens implies an internalized understanding of this difference. Note that now, the tokens \texttt{CC} and \texttt{C}, both inside the sidechain, report higher similarity (0.8) than that of Figure~\ref{fig:token_similarity}a. Overall, although the tokenizer in Llama3 may not perfectly differentiate certain aspects of polymer PSMILES, our approach to refining the tokens shows that the Llama3 can still effectively capture both the syntactic structure and the underlying chemical relationships within the PSMILES strings. While token-level embeddings retain considerably more information than mean-pooled embeddings, we opted for the aggregated embeddings for their simplicity and to demonstrate the effectiveness of the LLM embedding model in property prediction tasks.

\subsubsection{Performance of PolyLLMem}

\begin{table}[htbp]
    \centering

    \label{tab:polymer_property_performance}
    \resizebox{\textwidth}{!}{%
    \begin{tabular}{lccccccc}
    \hline
    \textbf{Property} & \textbf{PolyLLMem} & \textbf{LLM+MLP} & \textbf{Uni-Mol+MLP} & \textbf{LLM+XGB} & \textbf{Uni-Mol+XGB} & \textbf{MF+XGB} & \textbf{descriptors+XGB} \\
    \hline
    $\rho$       & $\bm{0.82 \pm 0.03 \uparrow}$ & $0.70 \pm 0.06$ & $0.74 \pm 0.05$ & $0.58 \pm 0.02$ & $0.67 \pm 0.03$ & $0.62 \pm 0.02$ & $0.73 \pm 0.05$ \\
    $T_g$        & $\bm{0.89 \pm 0.01 \uparrow}$ & $0.88 \pm 0.01$ & $0.85 \pm 0.01$ & $0.84 \pm 0.00$ & $0.82 \pm 0.01$ & $0.87 \pm 0.00$ & $0.87 \pm 0.00$ \\
    $T_m$        & $\bm{0.76 \pm 0.01 \uparrow}$ & $0.75 \pm 0.02$ & $0.70 \pm 0.01$ & $0.70 \pm 0.01$ & $0.63 \pm 0.01$ & $0.75 \pm 0.01$ & $0.68 \pm 0.02$ \\
    $T_d$        & $\bm{0.73 \pm 0.01 \uparrow}$          & $0.66 \pm 0.01$ & $0.63 \pm 0.04$ & $0.66 \pm 0.02$ & $0.59 \pm 0.01$ & $0.72 \pm 0.01$ & $0.71 \pm 0.01$ \\
    $\sigma_y$   & $0.56 \pm 0.12$         & $0.1 \pm 0.43$& $-0.43 \pm 0.32$ & $-0.40 \pm 0.82$ & $-0.83 \pm 1.51$ & $\bm{0.60 \pm 0.17 \uparrow}$& $0.12 \pm 0.39$ \\
    $\sigma_b$   & $\bm{0.32 \pm 0.07 \uparrow}$          & $0.15 \pm 0.14$ & $0.15 \pm 0.09$ & $0.26 \pm 0.19$ & $0.29 \pm 0.18$ & $0.28 \pm 0.13$ & $0.23 \pm 0.13$ \\
    $\epsilon_b$ & $0.24 \pm 0.04$          & $0.19 \pm 0.08$ & $0.04 \pm 0.10$ & $0.32 \pm 0.05$ & $0.31 \pm 0.04$ & $\bm{0.34 \pm 0.04\uparrow}$ & $0.10 \pm 0.08$ \\
    $E$          & $\bm{0.52 \pm 0.06 \uparrow}$          & $0.37 \pm 0.05$ & $0.40 \pm 0.05$ & $0.43 \pm 0.07$ & $0.28 \pm 0.07$ & $0.46 \pm 0.03$ & $0.34 \pm 0.13$ \\
    $\sigma$     & $0.45 \pm 0.05$          & $0.35 \pm 0.04$ & $0.35 \pm 0.08$ & $0.40 \pm 0.04$ & $0.33 \pm 0.04$ & $0.44 \pm 0.03$ & $\bm{0.48 \pm 0.02 \uparrow}$ \\
    $E_{gc}$     & $\bm{0.92 \pm 0.01 \uparrow}$          & $0.88 \pm 0.01$ & $0.88 \pm 0.01$ & $0.81 \pm 0.01$ & $0.80 \pm 0.02$ & $0.86 \pm 0.01$ & $0.88 \pm 0.01$ \\
    $X_c$        & $0.40 \pm 0.03$          & $\bm{0.44 \pm 0.03 \uparrow}$ & $0.27 \pm 0.09$ & $0.37 \pm 0.06$ & $0.26 \pm 0.08$ & $0.28 \pm 0.08$ & $0.31 \pm 0.05$ \\
    $E_{gb}$     & $\bm{0.94 \pm 0.01 \uparrow}$          & $0.90 \pm 0.01$ & $0.93 \pm 0.02$ & $0.84 \pm 0.02$ & $0.85 \pm 0.03$ & $0.85 \pm 0.01$ & $0.91 \pm 0.01$ \\
    $E_{at}$     & $\bm{0.96 \pm 0.02 \uparrow}$          & $0.90 \pm 0.03$ & $0.90 \pm 0.02$ & $0.74 \pm 0.07$ & $0.80 \pm 0.02$ & $0.81 \pm 0.03$ & $0.90 \pm 0.02$ \\
    $E_{ea}$     & $\bm{0.92 \pm 0.01 \uparrow}$          & $0.86 \pm 0.02$ & $0.91 \pm 0.02$ & $0.63 \pm 0.07$ & $0.75 \pm 0.03$ & $0.83 \pm 0.02$ & $0.79 \pm 0.02$ \\
    $E_i$        & $\bm{0.81 \pm 0.04 \uparrow}$          & $0.76 \pm 0.05$ & $0.75 \pm 0.03$ & $0.70 \pm 0.05$ & $0.62 \pm 0.04$ & $0.76 \pm 0.03$ & $0.69 \pm 0.05$ \\
    $n_c$        & $\bm{0.83 \pm 0.01 \uparrow}$          & $0.72 \pm 0.11$ & $0.72 \pm 0.02$ & $0.69 \pm 0.03$ & $0.69 \pm 0.04$ & $0.69 \pm 0.06$ & $0.82 \pm 0.03$ \\
    $\mu_{CO_2}$ & $\bm{0.83 \pm 0.02 \uparrow}$          & $0.76 \pm 0.06$ & $0.63 \pm 0.12$ & $0.73 \pm 0.03$ & $0.64 \pm 0.05$ & $0.73 \pm 0.04$ & $0.74 \pm 0.03$ \\
    $\mu_{H_2}$  & $\bm{0.85 \pm 0.03 \uparrow}$          & $0.80 \pm 0.04$ & $0.81 \pm 0.03$ & $0.81 \pm 0.03$ & $0.66 \pm 0.05$ & $0.85 \pm 0.04$ & $0.79 \pm 0.03$ \\
    $\mu_{CH_4}$ & $\bm{0.87 \pm 0.03 \uparrow}$          & $0.79 \pm 0.01$ & $0.84 \pm 0.02$ & $0.78 \pm 0.05$ & $0.72 \pm 0.05$ & $0.81 \pm 0.03$ & $0.80 \pm 0.01$ \\
    $\mu_{He}$   & $\bm{0.81 \pm 0.02 \uparrow}$          & $0.76 \pm 0.04$ & $0.74 \pm 0.03$ & $0.77 \pm 0.04$ & $0.65 \pm 0.07$ & $0.77 \pm 0.01$ & $0.65 \pm 0.05$ \\
    $\mu_{N_2}$  & $\bm{0.79 \pm 0.01\uparrow}$          & $0.79 \pm 0.02$ & $0.68 \pm 0.08$ & $0.61 \pm 0.05$ & $0.67 \pm 0.02$ & $0.78 \pm 0.02$ & $0.70 \pm 0.03$ \\
    $\mu_{O_2}$  & $\bm{0.87 \pm 0.01 \uparrow}$          & $0.86 \pm 0.03$ & $0.77 \pm 0.06$ & $0.75 \pm 0.03$ & $0.66 \pm 0.04$ & $0.78 \pm 0.04$ & $0.69 \pm 0.04$ \\
    \hline
    \end{tabular}
    \caption{Comparison of predictive performance (mean $R^2$ scores ± standard deviation) across various polymer properties for different models. Results were obtained using five-fold cross-validation on test datasets. PolyLLMem refers to the multimodal model integrating LLM-generated text embeddings and Uni-Mol structural embeddings. LLM+XX and Uni-Mol+XX denote models utilizing embeddings from Llama3 or Uni-Mol, respectively, as input features for the indicated methods (MLP, XGB). MF+XGB and Descriptors+XGB denote models using RDKit molecular fingerprints (MF) or RDKit molecular descriptors as input features for XGB. Best-performing results per property are highlighted in bold with arrows. The properties of gas permeabilities ($\mu_{x}$), tensile strength at break ($\sigma_b$), tensile strength at yield ($\sigma_y$), tensile strength at yield ($\sigma_y$), elongation at break ($\epsilon_b$), Young’s modulus($E$) and conductivity ($\sigma$) were trained on log scale and the $R^2$ value for those properties were reported on this scale.}
    }
    \label{tab:polymer_res}
\end{table}

\begin{figure}[h]
\centering
\includegraphics[width=1\textwidth]{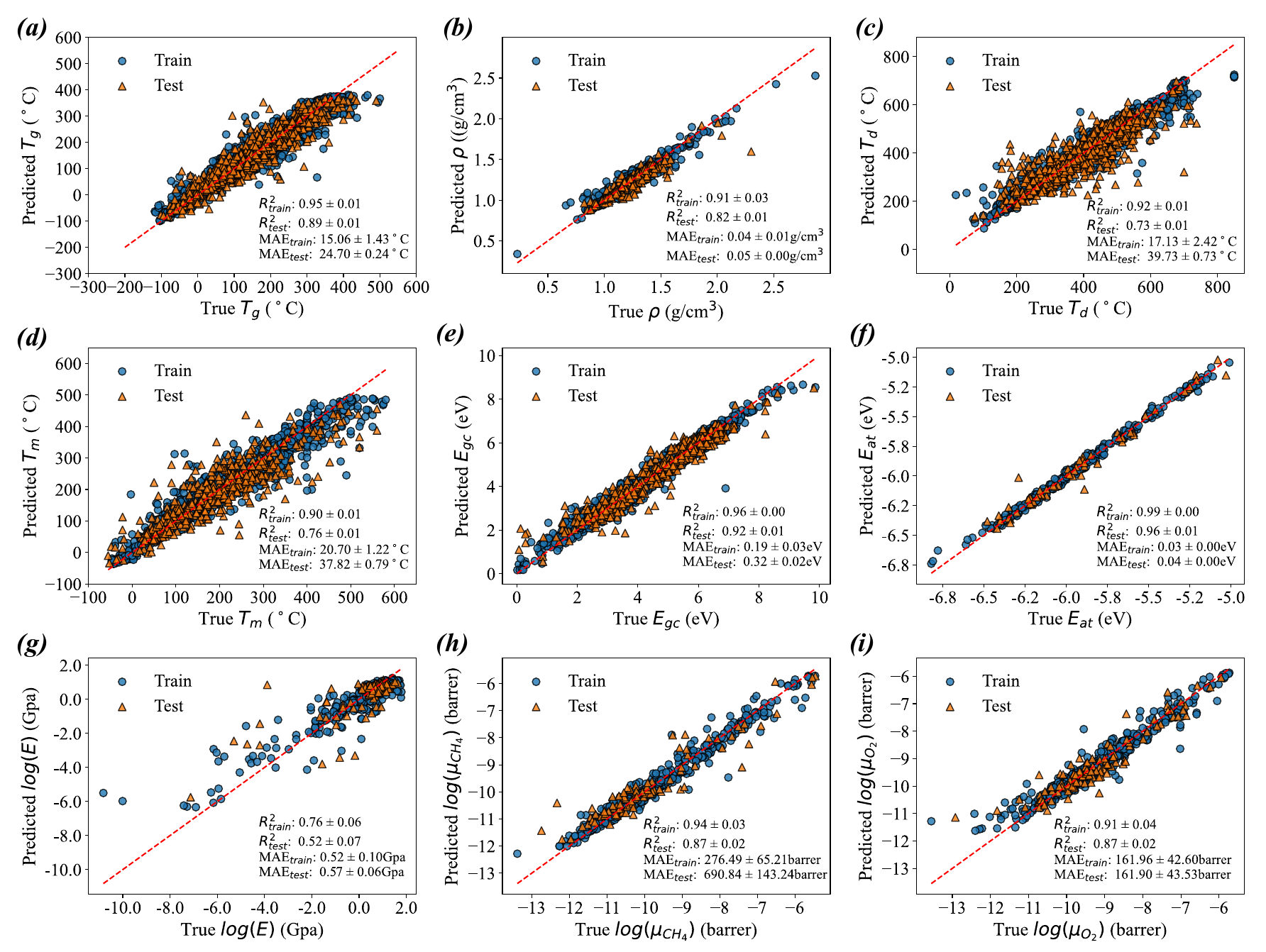}
\caption{Scatter plots of ground truth vs. predicted values for the selected properties: (a) $T_{g}$, (b) $\rho$, (c) $T_{d}$, (d) $T_{m}$, (e) $E_{gc}$, (f) $E_{at}$, (g) $E$, (h) $\mu_{CH_4}$, (i) $\mu_{O_2}$. The $R^2$ value for properties of $E$, $\mu_{CH_4}$, $\mu_{O_2}$ were calculated based on the training in the log scale,  whereas the MAE were reported on the original value.}
\label{fig4}
\end{figure}

After confirming that the LLM embeddings using Llama3 retained essential chemical information, we evaluated their predictive performance on over 22 distinct polymer properties. Initially, the generated 4096-dimensional embeddings for each polymer were used as input to a simple XGBoost model (LLM+XGB)\cite{chen2016xgboost}. The performance results for various polymer properties were summarized in Table~\ref{tab:polymer_data}, which reports the average $R^2$ values obtained from five-fold cross-validation on the test set. Despite straightforward, LLM+XGB already achieved satisfying performance on several polymer properties, including $T_{g}$, $E_{gc}$, $E_{gb}$, $\mu_{H_2}$, each exhibiting $R^2$ value above 0.8. These results support our UMAP analysis, indicating that the LLM embeddings capture meaningful chemical information and perform well on certain property prediction tasks. However, when compared with benchmark models trained on MF features (MF+XGB) or molecular descriptors (descriptors+XGB), LLM+XGB consistently underperformed across most property predictions, particularly when contrasted with MF+XGB. In addition, we evaluated other ML models such as RF, RR, AdaBoost, MLP, etc. using the LLM-generated embeddings. Among these, the model combining LLM embeddings with an MLP (LLM+MLP) demonstrated exceptional performance, yielding results that were comparable to those of the benchmark models on the majority of the property prediction tasks, as shown in Table~\ref{tab:polymer_data}. Additional results for other models were provided in the SI, as their performances were inferior to those of the benchmark models. We also trained MLP models using MF features and molecular descriptors respectively, but these results were inferior to those obtained with MF+XGB or descriptors+XGB (details were provided in the SI). 

Although the combination of LLM+MLP alone demonstrated exceptional performance overall, there remains room for improvement since the gains in prediction accuracy were modest and some property prediction tasks still underperformed compared to the baseline models. As LLM-based embeddings primarily capture textual information from PSMILES strings, crucial aspects of a molecule’s structure might be left out . To address this gap, we introduced Uni-Mol, a deep learning architecture that encodes the 3D geometry and conformational characteristics of small molecules, to generate embeddings that capture structural information. Integrating Uni-Mol-based embeddings with LLM-based ones,  our multimodal model PolyLLMem was formed. As shown in Table~\ref{tab:polymer_data} and Figure~\ref{fig4} , PolyLLMem demonstrated superior performance on most property prediction tasks compared to the baseline models, except for certain mechanical properties ($\sigma_y$,  $\epsilon_b$), crystallization tendency $X_c$ and conductivity $\sigma$. (Note that Uni-Mol embeddings alone as input features for ML models were also shown in Table~\ref{tab:polymer_data} as a comparison.)  Additional comparisons of prediction results of PolyLLMem with those of other state-of-the-art graph-based and transformer-based models, such as PolymerBERT, TransPolymer, and PolyGNN, as detailed in the SI. PolyLLMem exhibits performance that is comparable to, and in some cases exceeds, that of these benchmark models on the majority of polymer properties. Given the limited training data used in our study,  this highlights the model’s strong data efficiency and generalization capability. 


\subsubsection{Token-Level Interpretability}

\begin{figure}[h]
\centering
\includegraphics[width=0.9\textwidth]{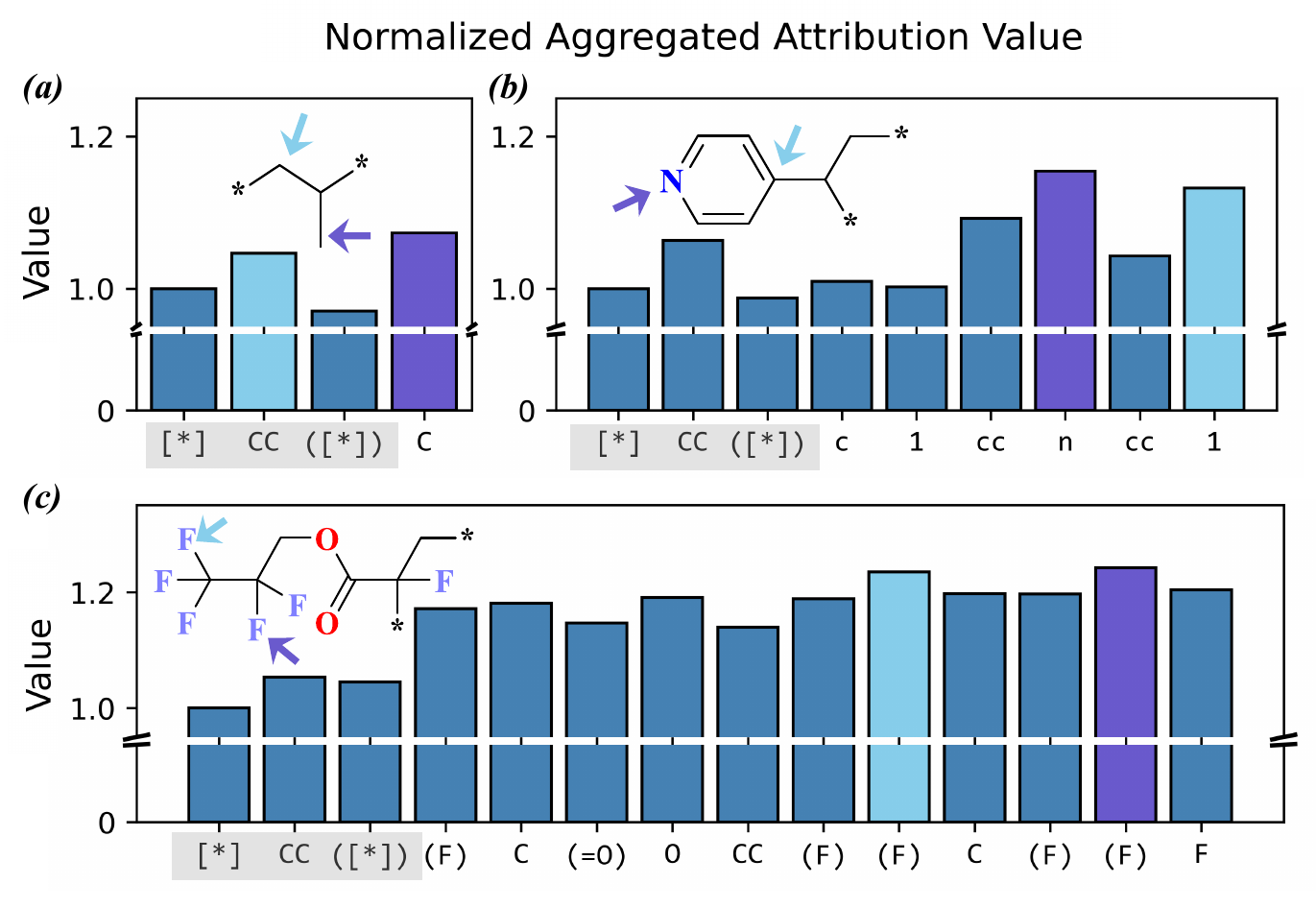}
\caption{Token-level attribution analysis for $T_g$ prediction using PolyLLMem of selected polymers: (a) \texttt{[*]CC([*])C}, (b) \texttt{[*]CC([*])\allowbreak c1ccncc1}, (c) \texttt{[*]CC([*])\allowbreak(F)C(=O)OCC(F)\allowbreak (F)C(F)(F)F}. Token-level attributions were computed using Integrated Gradients, highlighting the contribution of individual chemical tokens to model predictions. All values are normalized by the attribution value of token [*] for clarity.  Higher attribution values indicate greater significance in determining the predicted property. Light purple and light blue are used to tightly the top 2 influential tokens, with arrows of the same color notes for the corresponding substructure. The backbone of polymers is shaded in lightgrey.   }
\label{fig:attribute}
\end{figure}

Lastly, we commented on the interpretability of the model. The token-level embeddings were extracted using Llama3 tokenizer, following by appling a mean pooling layer to obtain the aggregated representation used by PolyLLMem. Using this setup, the Integrated Gradients was computed \cite{sundararajan2017axiomatic} along the path from a zero (baseline) input to the actual token-level embeddings. A wrapper function applied mean pooling over the embeddings, enabling the attribution scores to be assigned at the token level.  Such value can be correlated to the token's contribution to property prediction. As noted in the previous section regarding challenges with accurately splitting chemical notations, once we obtained the attribution values for each token, we applied the same merging strategy described earlier to consolidate and refine these values according to the refined token representations. Figure~\ref{fig:attribute} shows the refined token attributions for $T_{g}$ prediction from the selected polymers. For polymer \texttt{[*]CC([*])C} in Figure~\ref{fig:attribute}a, the tokens representing the carbon backbone (\texttt{CC}) and the sidechain carbon (\texttt{C}) show relatively high aggregated attribution values, indicating that the model emphasizes the alkyl nature. In Figure~\ref{fig:attribute}b, for polymer \texttt{[*]CC([*])c1ccncc1}, the tokens corresponding to the nitrogen-containing ring (\texttt{n}), and the ring closure marker (\texttt{1}) exhibit higher aggregated attribution values, showing the importance of the substituent effect in the benzene. This is as expected, as in polymer chemistry, heteroaromatic rings, particularly those containing nitrogen, can significantly affect chain packing and polarity, which in turn affects $T_{g}$. Lastly, in polymer \texttt{[*]CC([*])(F)C(=O)OCC(F)(F)C(F)(F)F}, which exhibits one of the highest $T_{g}$ values in our dataset, the tokens corresponding to the fluorine in the fluoromethyl \texttt{(including both –CF$_3$ and -CF$_2$-)} groups have the highest aggregated attribution values. This also aligns with real-world polymer chemistry, where fluorinated groups can significantly affect chain rigidity and consequently increase $T_{g}$. Collectively, the model’s attention to these functional groups and structural moieties demonstrates its ability to capture the key chemical features that determine $T_{g}$ in practice.

\newpage
\section{Conclusion}
In this study, we introduced a simple, lightweight, yet effective multimodal framework, PolyLLMem, which synergistically integrates LLM-based text embeddings with 3D molecular structure embeddings from Uni-Mol to predict a wide range of polymer properties. By leveraging PSMILES strings as a source of rich textual information and integrating them with geometrical descriptors, our approach successfully encapsulates both the chemical context inherent in polymer notations and the critical conformational features of polymer molecules. Our extensive evaluation across 22 distinct property prediction tasks demonstrated that PolyLLMem achieves competitive performance compared to established transformer and graph-based models that are specifically designed for the polymer domain without the requirements of  millions of data points or data augmentation. Notably, the LLM (Llama3) embeddings model alone (LLM+XGB) showed promise for properties such as $T_g$, $E_{gc}$, and $E_{gb}$, indicating some of the chemical information within polymer domains was already embedded in the Llama3. Further integration of the LLM embeddings model with 3D structural information enhanced predictive accuracy across most tasks. Additionally, using Integrated Gradients also revealed that the model effectively identifies key chemical motifs and structural features in line with established chemical intuition.
 
Despite these promising results, challenges remain, particularly in predicting certain mechanical properties. These discrepancies highlight the inherent complexities of polymer data and indicate the need for further refinement of data augmentation techniques and model architectures. Future work shall focus on enhancing the embedding model's complexity by leveraging token-level embeddings in conjunction with multi-head attention mechanisms, which may further improve the accuracy of polymer property predictions. Nonetheless, our work here already underscores the potential of combining LLM embeddings with molecular structure information to accelerate the discovery and optimization of polymer materials. 

\newpage

\subsection{Code availability}
Source code is available for academic use at https://github.com/zhangtr10\allowbreak/PolyLLMem.

\subsection{Conflict of Interest}
The authors have no conflicts to disclose.

\begin{acknowledgement}
This work employed Jetstream2 through allocation TG-MAT250013 from the Advanced Cyberinfrastructure Coordination Ecosystem, Services \& Support (ACCESS) and Delaware Advanced Research Workforce and Innovation Network (DARWIN).

\end{acknowledgement}

\bibliography{citation}

\begin{suppinfo}

\begin{figure}[htbp]
    \centering
    \includegraphics[width=\textwidth]{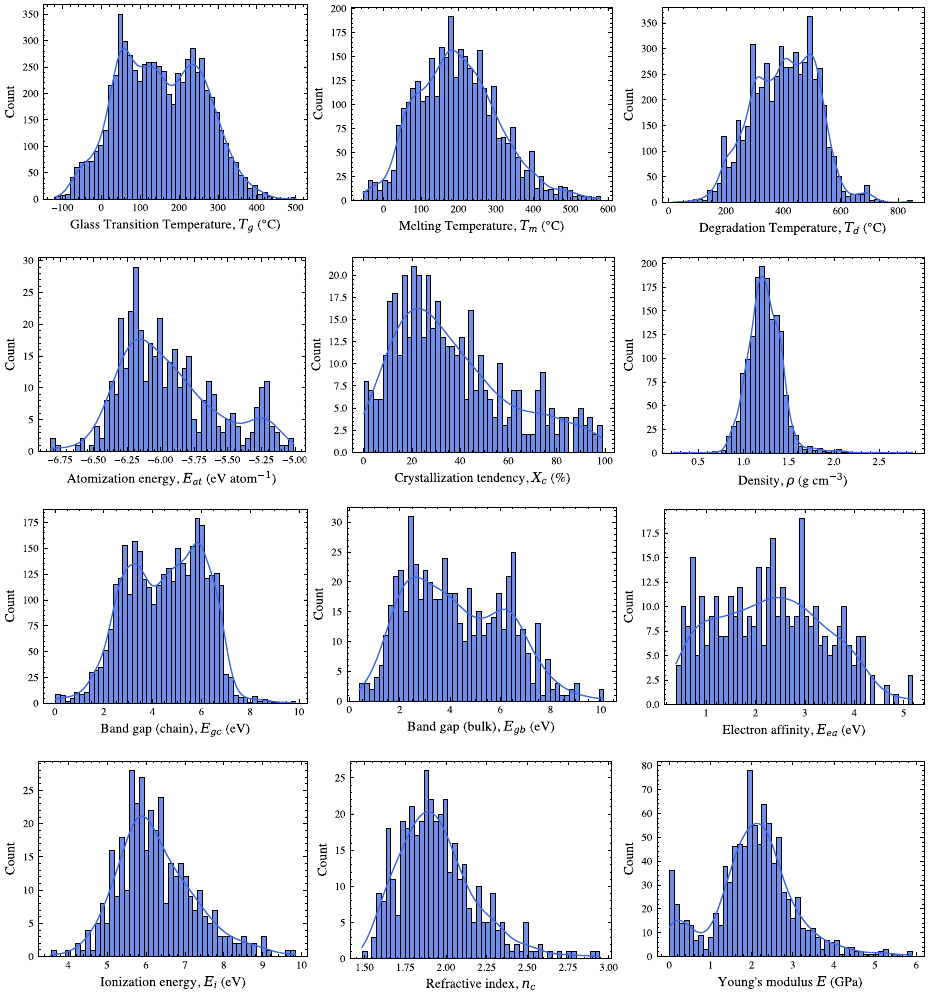}
    \label{fig:s1a}
\end{figure}

\clearpage

\begin{figure}[htbp]
    \centering
    \includegraphics[width=\textwidth]{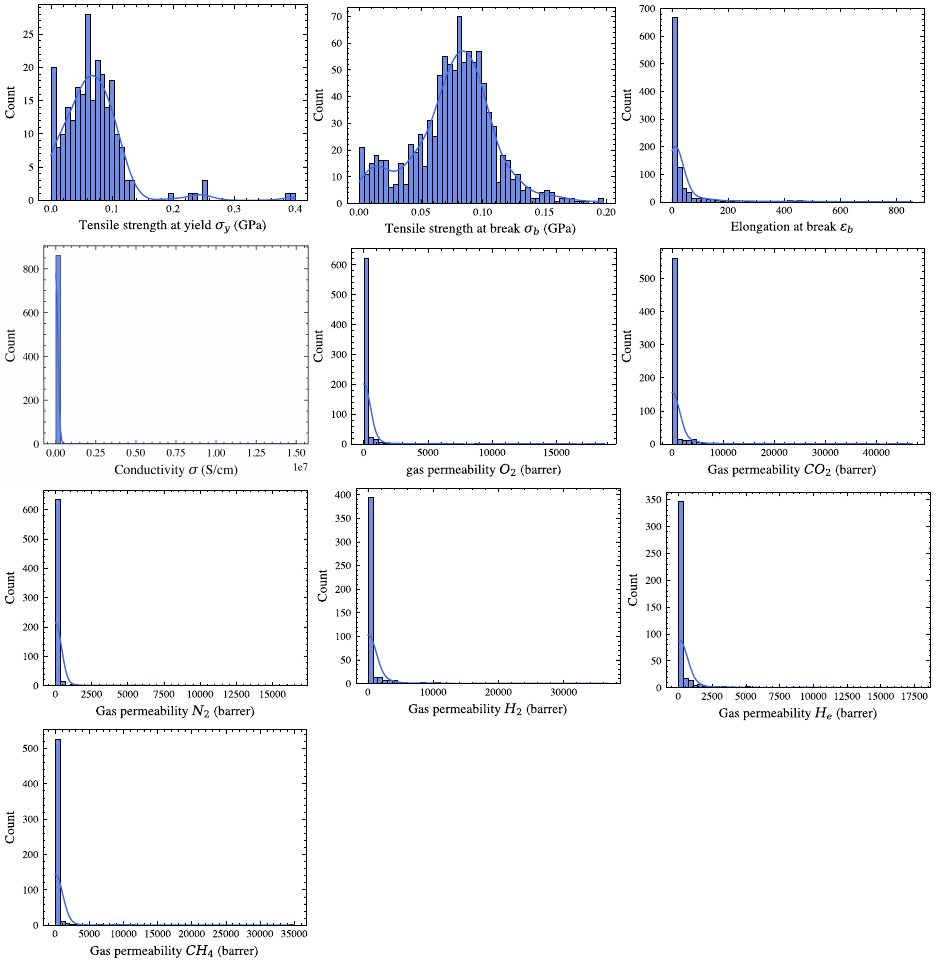}
    \caption*{\textbf{Figure S1.} The detailed distribution range for each property.}
    \label{fig:s1b}
\end{figure}

\clearpage
\begin{table}[htbp]
\centering
\caption*{\textbf{Table S1.}Comparison of predictive performance ($R^2$ scores ± standard deviation) for polymer properties across various machine learning models using the embeddings generated from Llama3 as input features. 

Methods used: RF (Random Forest), LR (Linear Regression), SVR (Support Vector Regression), FT (FastTree), RR (Ridge Regression), AdaBoost (Adaptive Boosting), GB (Gradient Boosting), and XGBoost (Extreme Gradient Boosting). }
\label{tab:ml_comparison}
\resizebox{\textwidth}{!}{%
\begin{tabular}{lcccccccc}
\toprule
\textbf{Property} & \textbf{RF}& \textbf{LR}& \textbf{SVR} & \textbf{FT}& \textbf{RR}& \textbf{AdaBoost} & \textbf{GB}& \textbf{XGBoost} \\
\midrule
$\rho$ & 0.51 ± 0.02 & 0.60 ± 0.06 & 0.61 ± 0.01 & 0.27 ± 0.13 & 0.65 ± 0.05 & 0.43 ± 0.02 & 0.60 ± 0.02 & 0.58 ± 0.02 \\
$T_g$ & 0.82 ± 0.00 & 0.77 ± 0.02 & 0.63 ± 0.00 & 0.63 ± 0.02 & 0.76 ± 0.02 & 0.73 ± 0.00 & 0.82 ± 0.00 & 0.84 ± 0.00 \\
$T_m$ & 0.67 ± 0.02 & 0.37 ± 0.05 & 0.30 ± 0.00 & 0.37 ± 0.09 & 0.52 ± 0.03 & 0.60 ± 0.01 & 0.70 ± 0.01 & 0.70 ± 0.01 \\
$T_d$ & 0.61 ± 0.01 & 0.20 ± 0.02 & 0.42 ± 0.00 & 0.28 ± 0.05 & 0.32 ± 0.03 & 0.51 ± 0.01 & 0.60 ± 0.01 & 0.66 ± 0.02 \\
$\sigma_y$ & -0.18 ± 0.45 & -3.50 ± 1.54 & 0.35 ± 0.01 & -5.20 ± 2.72 & -1.99 ± 0.86 & 0.52 ± 0.04 & 0.02 ± 0.38 & -5.01 ± 0.36 \\
$\sigma_b$ & 0.32 ± 0.08 & -0.67 ± 0.13 & 0.23 ± 0.01 & -0.24 ± 0.57 & 0.06 ± 0.12 & 0.25 ± 0.07 & 0.19 ± 0.27 & 0.26 ± 0.19 \\
$\epsilon_b$ & 0.36 ± 0.06 & -0.69 ± 0.33 & 0.33 ± 0.01 & -0.44 ± 0.06 & 0.06 ± 0.12 & 0.33 ± 0.03 & 0.39 ± 0.04 & 0.32 ± 0.05 \\
$E$ & 0.45 ± 0.09 & -0.59 ± 0.21 & 0.22 ± 0.03 & -0.51 ± 0.29 & 0.06 ± 0.14 & 0.43 ± 0.03 & 0.45 ± 0.05 & 0.43 ± 0.07 \\
$\sigma$ & 0.37 ± 0.05 & 0.08 ± 0.12 & 0.09 ± 0.01 & 0.09 ± 0.11 & 0.28 ± 0.07 & 0.38 ± 0.01 & 0.47 ± 0.01 & 0.40 ± 0.04 \\
$E_{gc}$ & 0.78 ± 0.01 & 0.70 ± 0.02 & 0.85 ± 0.00 & 0.59 ± 0.02 & 0.76 ± 0.02 & 0.73 ± 0.01 & 0.80 ± 0.00 & 0.81 ± 0.01 \\
$X_c$ & 0.39 ± 0.04 & -1.06 ± 0.31 & 0.00 ± 0.02 & -0.12 ± 0.17 & -0.48 ± 0.08 & 0.44 ± 0.03 & 0.45 ± 0.03 & 0.37 ± 0.06 \\
$E_{gb}$ & 0.81 ± 0.02 & 0.82 ± 0.02 & 0.85 ± 0.01 & 0.49 ± 0.09 & 0.86 ± 0.02 & 0.84 ± 0.01 & 0.86 ± 0.01 & 0.84 ± 0.02 \\
$E_{at}$ & 0.72 ± 0.04 & 0.90 ± 0.03 & 0.83 ± 0.03 & 0.50 ± 0.06 & 0.90 ± 0.03 & 0.78 ± 0.02 & 0.81 ± 0.02 & 0.74 ± 0.07 \\
$E_{ea}$ & 0.66 ± 0.05 & 0.68 ± 0.13 & 0.79 ± 0.01 & 0.43 ± 0.13 & 0.78 ± 0.07 & 0.75 ± 0.02 & 0.77 ± 0.04 & 0.63 ± 0.07 \\
$E_i$ & 0.73 ± 0.02 & 0.55 ± 0.05 & 0.80 ± 0.02 & 0.33 ± 0.11 & 0.61 ± 0.03 & 0.72 ± 0.02 & 0.77 ± 0.03 & 0.70 ± 0.05 \\
$n_c$ & 0.70 ± 0.02 & 0.38 ± 0.28 & 0.71 ± 0.04 & 0.49 ± 0.07 & 0.60 ± 0.13 & 0.72 ± 0.03 & 0.71 ± 0.04 & 0.69 ± 0.03 \\
$\mu_{CO_2}$ & 0.71 ± 0.02 & 0.76 ± 0.02 & 0.81 ± 0.01 & 0.41 ± 0.11 & 0.79 ± 0.02 & 0.68 ± 0.02 & 0.78 ± 0.02 & 0.73 ± 0.03 \\
$\mu_{H_2}$ & 0.79 ± 0.01 & 0.79 ± 0.03 & 0.84 ± 0.01 & 0.56 ± 0.14 & 0.81 ± 0.03 & 0.81 ± 0.01 & 0.84 ± 0.03 & 0.81 ± 0.03 \\
$\mu_{CH_4}$ & 0.74 ± 0.06 & 0.67 ± 0.05 & 0.82 ± 0.01 & 0.63 ± 0.07 & 0.70 ± 0.05 & 0.76 ± 0.01 & 0.82 ± 0.02 & 0.78 ± 0.05 \\
$\mu_{He}$ & 0.76 ± 0.01 & 0.76 ± 0.04 & 0.80 ± 0.02 & 0.46 ± 0.17 & 0.77 ± 0.04 & 0.78 ± 0.03 & 0.80 ± 0.03 & 0.77 ± 0.04 \\
$\mu_{N_2}$ & 0.66 ± 0.07 & 0.71 ± 0.02 & 0.77 ± 0.01 & 0.20 ± 0.09 & 0.74 ± 0.01 & 0.74 ± 0.02 & 0.75 ± 0.03 & 0.61 ± 0.05 \\
$\mu_{O_2}$ & 0.72 ± 0.02 & 0.78 ± 0.03 & 0.82 ± 0.00 & 0.46 ± 0.06 & 0.82 ± 0.02 & 0.71 ± 0.01 & 0.80 ± 0.02 & 0.75 ± 0.03 \\
\bottomrule
\end{tabular}%
}
\end{table}

\begin{table}[htbp]
\centering
\caption*{\textbf{Table S2.} Comparison of predictive performance ($R^2$ scores ± standard deviation) for polymer properties across various machine learning models using the embeddings generated from Uni-Mol as input features.}
\label{tab:ml_models_results}
\resizebox{\textwidth}{!}{%
\begin{tabular}{lcccccccc}
\toprule
\textbf{Property} & \textbf{RF}& \textbf{LR}& \textbf{SVR} & \textbf{DT}& \textbf{RR}& \textbf{AdaBoost} & \textbf{GB}& \textbf{XGBoost} \\
\midrule
$\rho$ & 0.62 ± 0.01 & 0.62 ± 0.03 & 0.66 ± 0.01 & 0.19 ± 0.17 & 0.67 ± 0.03 & 0.57 ± 0.02 & 0.72 ± 0.03 & 0.67 ± 0.03 \\
$T_g$ & 0.80 ± 0.00 & 0.81 ± 0.01 & 0.66 ± 0.00 & 0.57 ± 0.02 & 0.82 ± 0.01 & 0.74 ± 0.01 & 0.82 ± 0.00 & 0.82 ± 0.01 \\
$T_m$ & 0.60 ± 0.01 & 0.26 ± 0.03 & 0.28 ± 0.00 & 0.17 ± 0.03 & 0.35 ± 0.02 & 0.57 ± 0.01 & 0.65 ± 0.00 & 0.63 ± 0.01 \\
$T_d$ & 0.56 ± 0.01 & 0.46 ± 0.02 & 0.38 ± 0.00 & 0.14 ± 0.04 & 0.47 ± 0.02 & 0.49 ± 0.01 & 0.59 ± 0.01 & 0.59 ± 0.01 \\
$\sigma_y$ & -0.22 ± 0.87 & -0.02 ± 0.15 & 0.19 ± 0.01 & -3.83 ± 6.32 & -0.01 ± 0.15 & 0.42 ± 0.12 & 0.13 ± 0.56 & -1.45 ± 3.53 \\
$\sigma_b$ & 0.30 ± 0.05 & -0.09 ± 0.11 & 0.20 ± 0.01 & -0.44 ± 0.39 & -0.08 ± 0.11 & 0.34 ± 0.05 & 0.32 ± 0.11 & 0.29 ± 0.18 \\
$\epsilon_b$ & 0.32 ± 0.05 & -0.45 ± 0.09 & 0.22 ± 0.01 & -0.25 ± 0.07 & -0.41 ± 0.08 & 0.32 ± 0.03 & 0.40 ± 0.01 & 0.31 ± 0.04 \\
$E$ & 0.30 ± 0.08 & -0.04 ± 0.03 & 0.12 ± 0.01 & -0.06 ± 0.26 & -0.02 ± 0.02 & 0.29 ± 0.06 & 0.39 ± 0.04 & 0.28 ± 0.07 \\
$\sigma$ & 0.32 ± 0.04 & 0.06 ± 0.12 & 0.02 ± 0.00 & -0.15 ± 0.10 & 0.07 ± 0.11 & 0.35 ± 0.05 & 0.39 ± 0.04 & 0.33 ± 0.04 \\
$E_{gc}$ & 0.77 ± 0.01 & 0.64 ± 0.01 & 0.86 ± 0.00 & 0.54 ± 0.05 & 0.68 ± 0.01 & 0.75 ± 0.00 & 0.82 ± 0.00 & 0.80 ± 0.02 \\
$X_c$ & 0.27 ± 0.02 & 0.07 ± 0.07 & -0.01 ± 0.01 & -0.47 ± 0.34 & 0.08 ± 0.07 & 0.30 ± 0.05 & 0.31 ± 0.03 & 0.26 ± 0.08 \\
$E_{gb}$ & 0.84 ± 0.03 & 0.90 ± 0.01 & 0.89 ± 0.01 & 0.65 ± 0.06 & 0.90 ± 0.01 & 0.87 ± 0.01 & 0.89 ± 0.01 & 0.85 ± 0.03 \\
$E_{at}$ & 0.75 ± 0.03 & 0.94 ± 0.01 & 0.81 ± 0.03 & 0.62 ± 0.06 & 0.94 ± 0.01 & 0.77 ± 0.03 & 0.85 ± 0.02 & 0.80 ± 0.02 \\
$E_{ea}$ & 0.70 ± 0.03 & 0.91 ± 0.01 & 0.83 ± 0.02 & 0.49 ± 0.12 & 0.91 ± 0.01 & 0.79 ± 0.02 & 0.84 ± 0.02 & 0.75 ± 0.03 \\
$E_i$ & 0.68 ± 0.02 & 0.74 ± 0.02 & 0.80 ± 0.01 & 0.30 ± 0.11 & 0.74 ± 0.02 & 0.70 ± 0.03 & 0.72 ± 0.04 & 0.62 ± 0.04 \\
$n_c$ & 0.71 ± 0.03 & 0.72 ± 0.05 & 0.69 ± 0.03 & 0.52 ± 0.10 & 0.72 ± 0.05 & 0.72 ± 0.04 & 0.74 ± 0.02 & 0.69 ± 0.04 \\
$\mu_{CO_2}$ & 0.61 ± 0.02 & 0.65 ± 0.05 & 0.79 ± 0.01 & 0.28 ± 0.18 & 0.66 ± 0.05 & 0.64 ± 0.02 & 0.69 ± 0.02 & 0.64 ± 0.05 \\
$\mu_{H_2}$ & 0.65 ± 0.05 & 0.77 ± 0.03 & 0.79 ± 0.01 & 0.41 ± 0.09 & 0.78 ± 0.03 & 0.69 ± 0.03 & 0.73 ± 0.03 & 0.66 ± 0.05 \\
$\mu_{CH_4}$ & 0.66 ± 0.04 & 0.73 ± 0.07 & 0.79 ± 0.01 & 0.36 ± 0.17 & 0.74 ± 0.07 & 0.76 ± 0.02 & 0.77 ± 0.01 & 0.72 ± 0.05 \\
$\mu_{He}$ & 0.66 ± 0.02 & 0.74 ± 0.04 & 0.75 ± 0.02 & 0.43 ± 0.10 & 0.78 ± 0.02 & 0.65 ± 0.03 & 0.70 ± 0.02 & 0.65 ± 0.07 \\
$\mu_{N_2}$ & 0.64 ± 0.02 & 0.68 ± 0.02 & 0.75 ± 0.00 & 0.34 ± 0.10 & 0.69 ± 0.02 & 0.69 ± 0.01 & 0.69 ± 0.03 & 0.67 ± 0.02 \\
$\mu_{O_2}$ & 0.68 ± 0.03 & 0.70 ± 0.04 & 0.78 ± 0.01 & 0.42 ± 0.08 & 0.71 ± 0.03 & 0.68 ± 0.02 & 0.74 ± 0.02 & 0.66 ± 0.04 \\
\bottomrule
\end{tabular}%
}
\end{table}

\begin{table}[htbp]
\centering
\caption*{\textbf{Table S3.} Comparison of predictive performance ($R^2$ scores ± standard deviation) for polymer properties across various machine learning models using Morgan Fingerprint as input features.}
\label{tab:ml_models_mlp}
\resizebox{\textwidth}{!}{%
\begin{tabular}{lccccccccc}
\toprule
\textbf{Property} & \textbf{RF}& \textbf{LR}& \textbf{SVR} & \textbf{DT}& \textbf{RR}& \textbf{AdaBoost} & \textbf{GB}& \textbf{XGBoost} & \textbf{MLP} \\
\midrule
$\rho$ & 0.57 ± 0.04 & -2.51 ± 1.10 & 0.41 ± 0.01 & 0.42 ± 0.08 & -0.49 ± 0.25 & 0.39 ± 0.04 & 0.54 ± 0.01 & 0.62 ± 0.02 & 0.38 ± 0.04 \\
$T_g$ & 0.86 ± 0.00 & 0.79 ± 0.01 & 0.35 ± 0.00 & 0.75 ± 0.00 & 0.79 ± 0.01 & 0.68 ± 0.01 & 0.81 ± 0.00 & 0.87 ± 0.00 & 0.85 ± 0.00 \\
$T_m$ & 0.73 ± 0.01 & 0.22 ± 0.09 & 0.11 ± 0.00 & 0.54 ± 0.03 & 0.24 ± 0.08 & 0.50 ± 0.02 & 0.67 ± 0.01 & 0.75 ± 0.01 & 0.71 ± 0.01 \\
$T_d$ & 0.68 ± 0.02 & 0.54 ± 0.01 & 0.19 ± 0.00 & 0.47 ± 0.02 & 0.54 ± 0.01 & 0.46 ± 0.02 & 0.60 ± 0.00 & 0.72 ± 0.01 & 0.64 ± 0.02 \\
$\sigma_y$ & 0.41 ± 0.53 & -0.82 ± 0.60 & 0.52 ± 0.02 & 0.58 ± 0.13 & 0.22 ± 0.24 & 0.42 ± 0.05 & 0.72 ± 0.06 & 0.62 ± 0.17 & 0.27 ± 0.10 \\
$\sigma_b$ & 0.27 ± 0.12 & -1.07 ± 0.78 & 0.32 ± 0.02 & -0.43 ± 0.04 & 0.43 ± 0.12 & -0.24 ± 0.20 & 0.37 ± 0.10 & 0.28 ± 0.13 & 0.21 ± 0.05 \\
$\epsilon_b$ & 0.33 ± 0.03 & -4.21 ± 2.33 & 0.45 ± 0.01 & 0.00 ± 0.04 & 0.19 ± 0.04 & 0.26 ± 0.04 & 0.40 ± 0.03 & 0.34 ± 0.04 & 0.29 ± 0.04 \\
$E$ & 0.64 ± 0.04 & -1.00 ± 0.89 & 0.40 ± 0.02 & 0.32 ± 0.19 & 0.55 ± 0.07 & 0.16 ± 0.28 & 0.53 ± 0.06 & 0.46 ± 0.03 & 0.37 ± 0.05 \\
$\sigma$ & 0.38 ± 0.04 & -0.22 ± 0.25 & 0.18 ± 0.00 & -0.01 ± 0.04 & 0.29 ± 0.10 & 0.20 ± 0.01 & 0.39 ± 0.04 & 0.43 ± 0.04 & 0.60 ± 0.01 \\
$E_{gc}$ & 0.85 ± 0.01 & 0.67 ± 0.02 & 0.78 ± 0.00 & 0.74 ± 0.02 & 0.67 ± 0.02 & 0.69 ± 0.01 & 0.82 ± 0.01 & 0.86 ± 0.01 & 0.82 ± 0.01 \\
$X_c$ & 0.39 ± 0.06 & -0.12 ± 0.25 & -0.04 ± 0.01 & -0.02 ± 0.12 & -0.03 ± 0.20 & 0.30 ± 0.02 & 0.34 ± 0.05 & 0.28 ± 0.08 & 0.42 ± 0.04 \\
$E_{gb}$ & 0.85 ± 0.02 & 0.66 ± 0.03 & 0.55 ± 0.01 & 0.72 ± 0.06 & 0.68 ± 0.03 & 0.79 ± 0.02 & 0.86 ± 0.01 & 0.85 ± 0.01 & 0.76 ± 0.03 \\
$E_{at}$ & 0.75 ± 0.03 & 0.72 ± 0.04 & 0.32 ± 0.03 & 0.65 ± 0.03 & 0.71 ± 0.05 & 0.73 ± 0.02 & 0.82 ± 0.02 & 0.81 ± 0.03 & -4.57 ± 1.33 \\
$E_{ea}$ & 0.82 ± 0.02 & 0.61 ± 0.03 & 0.40 ± 0.02 & 0.75 ± 0.05 & 0.62 ± 0.04 & 0.76 ± 0.02 & 0.83 ± 0.02 & 0.83 ± 0.02 & 0.79 ± 0.01 \\
$E_i$ & 0.73 ± 0.02 & 0.57 ± 0.06 & 0.52 ± 0.02 & 0.56 ± 0.09 & 0.62 ± 0.06 & 0.67 ± 0.02 & 0.77 ± 0.01 & 0.76 ± 0.03 & -0.01 ± 0.04 \\
$n_c$ & 0.65 ± 0.05 & 0.21 ± 0.07 & 0.46 ± 0.02 & 0.42 ± 0.11 & 0.47 ± 0.10 & 0.66 ± 0.03 & 0.70 ± 0.03 & 0.69 ± 0.06 & 0.36 ± 0.17 \\
$\mu_{CO_2}$ & 0.74 ± 0.02 & 0.35 ± 0.17 & 0.61 ± 0.01 & 0.49 ± 0.12 & 0.52 ± 0.05 & 0.65 ± 0.01 & 0.76 ± 0.01 & 0.73 ± 0.04 & 0.29 ± 0.07 \\
$\mu_{H_2}$ & 0.86 ± 0.01 & 0.65 ± 0.07 & 0.64 ± 0.02 & 0.72 ± 0.06 & 0.68 ± 0.05 & 0.77 ± 0.02 & 0.82 ± 0.02 & 0.85 ± 0.04 & 0.18 ± 0.09 \\
$\mu_{CH_4}$ & 0.79 ± 0.01 & 0.48 ± 0.09 & 0.56 ± 0.01 & 0.73 ± 0.04 & 0.58 ± 0.08 & 0.71 ± 0.01 & 0.81 ± 0.02 & 0.81 ± 0.03 & 0.52 ± 0.02 \\
$\mu_{He}$ & 0.79 ± 0.01 & 0.42 ± 0.13 & 0.60 ± 0.02 & 0.67 ± 0.05 & 0.57 ± 0.06 & 0.77 ± 0.02 & 0.80 ± 0.01 & 0.77 ± 0.01 & -0.47 ± 0.19 \\
$\mu_{N_2}$ & 0.74 ± 0.01 & 0.35 ± 0.10 & 0.61 ± 0.01 & 0.63 ± 0.06 & 0.54 ± 0.03 & 0.66 ± 0.01 & 0.76 ± 0.01 & 0.78 ± 0.02 & 0.12 ± 0.09 \\
$\mu_{O_2}$ & 0.71 ± 0.03 & 0.44 ± 0.09 & 0.64 ± 0.01 & 0.56 ± 0.03 & 0.61 ± 0.04 & 0.65 ± 0.01 & 0.75 ± 0.02 & 0.78 ± 0.04 & 0.43 ± 0.02 \\
\bottomrule
\end{tabular}%
}
\end{table}

\begin{table}[htbp]
\centering
\caption*{\textbf{Table S4.} Comparison of predictive performance ($R^2$ scores ± standard deviation) for polymer properties across various machine learning models using the molecular descriptors as input features.}
\label{tab:ml_model_polymer_properties}
\resizebox{\textwidth}{!}{%
\begin{tabular}{lccccccc}
\toprule
\textbf{Property} & \textbf{RF} & \textbf{SVR} & \textbf{DT} & \textbf{AdaBoost} & \textbf{GB} & \textbf{XGBoost} & \textbf{MLP} \\
\midrule
$\rho$ & 0.66 ± 0.03 & 0.02 ± 0.01 & 0.61 ± 0.07 & 0.54 ± 0.02 & 0.72 ± 0.02 & 0.73 ± 0.05 & -0.06 ± 0.08 \\
$T_g$ & 0.85 ± 0.00 & 0.02 ± 0.00 & 0.71 ± 0.02 & 0.74 ± 0.00 & 0.83 ± 0.00 & 0.87 ± 0.00 & 0.87 ± 0.00 \\
$T_m$ & 0.64 ± 0.01 & 0.01 ± 0.00 & 0.42 ± 0.06 & 0.53 ± 0.01 & 0.66 ± 0.01 & 0.68 ± 0.02 & 0.69 ± 0.03 \\
$T_d$ & 0.69 ± 0.01 & 0.02 ± 0.00 & 0.41 ± 0.06 & 0.48 ± 0.01 & 0.63 ± 0.01 & 0.71 ± 0.01 & 0.64 ± 0.00 \\
$\sigma_y$ & 0.47 ± 0.26 & -0.03 ± 0.01 & 0.32 ± 0.34 & 0.56 ± 0.30 & 0.35 ± 0.34 & 0.12 ± 0.39 & 0.19 ± 0.11 \\
$\sigma_b$ & 0.45 ± 0.12 & -0.07 ± 0.00 & -0.03 ± 0.44 & 0.37 ± 0.16 & 0.45 ± 0.13 & 0.48 ± 0.02 & 0.27 ± 0.08 \\
$\epsilon_b$ & 0.17 ± 0.03 & -0.06 ± 0.00 & -0.28 ± 0.15 & 0.12 ± 0.04 & 0.14 ± 0.05 & 0.29 ± 0.13 & -0.04 ± 0.06 \\
$E$ & 0.40 ± 0.12 & -0.07 ± 0.01 & -0.39 ± 0.31 & 0.14 ± 0.13 & 0.35 ± 0.10 & 0.10 ± 0.08 & 0.38 ± 0.05 \\
$\sigma$ & 0.47 ± 0.05 & -0.17 ± 0.01 & 0.15 ± 0.10 & 0.36 ± 0.04 & 0.44 ± 0.02 & 0.39 ± 0.10 & 0.34 ± 0.10 \\
$E_{gc}$ & 0.86 ± 0.01 & 0.00 ± 0.00 & 0.75 ± 0.02 & 0.74 ± 0.01 & 0.85 ± 0.00 & 0.88 ± 0.01 & 0.83 ± 0.01 \\
$X_c$ & 0.39 ± 0.02 & -0.08 ± 0.02 & -0.00 ± 0.13 & 0.36 ± 0.02 & 0.39 ± 0.04 & 0.31 ± 0.05 & 0.49 ± 0.02 \\
$E_{gb}$ & 0.90 ± 0.01 & -0.06 ± 0.01 & 0.84 ± 0.01 & 0.88 ± 0.01 & 0.91 ± 0.01 & 0.91 ± 0.01 & 0.84 ± 0.04 \\
$E_{at}$ & 0.88 ± 0.04 & -0.00 ± 0.00 & 0.79 ± 0.08 & 0.86 ± 0.02 & 0.92 ± 0.03 & 0.90 ± 0.02 & -4.16 ± 0.81 \\
$E_{ea}$ & 0.80 ± 0.01 & -0.09 ± 0.04 & 0.68 ± 0.03 & 0.80 ± 0.01 & 0.84 ± 0.01 & 0.79 ± 0.02 & -25.96 ± 33.81 \\
$E_i$ & 0.73 ± 0.02 & -0.10 ± 0.04 & 0.42 ± 0.03 & 0.68 ± 0.02 & 0.72 ± 0.03 & 0.69 ± 0.05 & -52.32 ± 23.35 \\
$n_c$ & 0.78 ± 0.04 & -0.07 ± 0.02 & 0.58 ± 0.05 & 0.80 ± 0.03 & 0.83 ± 0.05 & 0.82 ± 0.03 & -0.25 ± 0.26 \\
$\mu_{CO_2}$ & 0.75 ± 0.03 & -0.06 ± 0.00 & 0.51 ± 0.09 & 0.64 ± 0.01 & 0.75 ± 0.02 & 0.74 ± 0.03 & 0.14 ± 0.18 \\
$\mu_{H_2}$ & 0.76 ± 0.03 & -0.03 ± 0.01 & 0.61 ± 0.09 & 0.73 ± 0.01 & 0.79 ± 0.02 & 0.79 ± 0.03 & 0.28 ± 0.07 \\
$\mu_{CH_4}$ & 0.79 ± 0.01 & -0.03 ± 0.01 & 0.72 ± 0.01 & 0.74 ± 0.02 & 0.79 ± 0.01 & 0.80 ± 0.01 & 0.29 ± 0.15 \\
$\mu_{He}$ & 0.70 ± 0.06 & -0.08 ± 0.02 & 0.49 ± 0.14 & 0.66 ± 0.05 & 0.69 ± 0.07 & 0.65 ± 0.05 & -0.02 ± 0.38 \\
$\mu_{N_2}$ & 0.71 ± 0.01 & -0.04 ± 0.01 & 0.48 ± 0.08 & 0.65 ± 0.01 & 0.73 ± 0.03 & 0.70 ± 0.03 & 0.32 ± 0.13 \\
$\mu_{O_2}$ & 0.69 ± 0.04 & -0.03 ± 0.01 & 0.44 ± 0.09 & 0.64 ± 0.03 & 0.71 ± 0.03 & 0.69 ± 0.04 & 0.61 ± 0.06 \\
\bottomrule
\end{tabular}%
}
\end{table}

\begin{table}[htbp]
    \centering
    \caption*{\textbf{Table S5.} Comparison of predictive performance (mean $R^2$ scores ± standard deviation) across various polymer properties for different models. The data for polyBERT\cite{kuenneth2023polybert},  Transpolymer\cite{xu2023transpolymer}, and single task (ST) polyGNN\cite{gurnani2023polymer} were obtained from their original paper.}
    \label{tab:polymer_property_performance}
    \resizebox{\textwidth}{!}{%
    \begin{tabular}{lcccc}
    \hline
    \textbf{Property} & \textbf{PolyLLMem} & \textbf{PolymerBERT} & \textbf{Transpolymer} & \textbf{ST polyGNN} \\
    \hline
    Train data & 0.02M & 100M & 5M & 0.02M with augmentation \\
    $\rho$ & $0.82 \pm 0.01$ & $0.75 \pm 0.03$ & - & $0.90 \pm 0.01$ \\
    $T_g$ & $0.89 \pm 0.01$ & $0.92 \pm 0.01$ & - & $0.89 \pm 0.01$ \\
    $T_m$ & $0.76 \pm 0.01$ & $0.84 \pm 0.02$ & - & $0.76 \pm 0.03$ \\
    $T_d$ & $0.73 \pm 0.01$ & $0.70 \pm 0.03$ & - & $0.66 \pm 0.02$ \\
    $\sigma_y$ & $0.56 \pm 0.12$ & $0.80 \pm 0.08$ & - & - \\
    $\sigma_b$ & $0.32 \pm 0.07$ & $0.76 \pm 0.05$ & - & $0.50 \pm 0.20$ \\
    $\epsilon_b$ & $0.24 \pm 0.04$ & $0.60 \pm 0.06$ & - & - \\
    $E$ & $0.52 \pm 0.06$ & $0.75 \pm 0.70$ & - & $0.43 \pm 0.20$ \\
    $\sigma$ & $0.45 \pm 0.05$ & - & - & - \\
    $E_{gc}$ & $0.92 \pm 0.01$ & $0.89 \pm 0.02$ & $0.92$ & $0.92 \pm 0.01$ \\
    $X_c$ & $0.40 \pm 0.03$ & $0.45 \pm 0.11$ & $0.50$ & $0.40 \pm 0.07$ \\
    $E_{gb}$ & $0.94 \pm 0.01$ & $0.93 \pm 0.01$ & $0.93$ & $0.84 \pm 0.07$ \\
    $E_{at}$ & $0.96 \pm 0.01$ & $0.85 \pm 0.02$ & - & $0.96 \pm 0.10$ \\
    $E_{ea}$ & $0.92 \pm 0.01$ & $0.93 \pm 0.03$ & $0.91$ & $0.78 \pm 0.10$ \\
    $E_i$ & $0.81 \pm 0.03$ & $0.82 \pm 0.07$ & $0.84$ & - \\
    $n_c$ & $0.83 \pm 0.01$ & $0.86 \pm 0.06$ & $0.82$ & $0.54 \pm 0.30$ \\
    $\mu_{CO_2}$ & $0.83 \pm 0.02$ & $0.94 \pm 0.02$ & - & $0.87 \pm 0.03$ \\
    $\mu_{H_2}$ & $0.85 \pm 0.03$ & $0.97 \pm 0.01$ & - & $0.91 \pm 0.02$ \\
    $\mu_{CH_4}$ & $0.87 \pm 0.03$ & $0.95 \pm 0.03$ & - & $0.90 \pm 0.03$ \\
    $\mu_{He}$ & $0.81 \pm 0.02$ & $0.95 \pm 0.02$ & - & $0.88 \pm 0.04$ \\
    $\mu_{N_2}$ & $0.79 \pm 0.01$ & $0.97 \pm 0.01$ & - & $0.83 \pm 0.10$ \\
    $\mu_{O_2}$ & $0.87 \pm 0.01$ & $0.96 \pm 0.01$ & - & $0.85 \pm 0.03$ \\
    \hline
    \end{tabular}
    }
\end{table}
\end{suppinfo}

\begin{table}[htbp]
\centering
\caption*{\textbf{Table S6.} Finetuning hyperparameters for PolyLLMem.}
\label{tab:finetune_hyperparams}
\begin{tabular}{ll}
\toprule
\textbf{Hyperparameter} & \textbf{Range} \\
\midrule
Batch size & \{8, 64\} \\
Hidden size & \{512, 4096\} \\
Rank & \{4, 32\} \\
Alpha & \{4, 128\} \\
Learning Rate & \{$5 \times 10^{-5}$, $1 \times 10^{-4}$\} \\
Weight Decay & \{0.001, 0.00001\} \\
Dropout rate& \{0.0, 0.5\} \\
\bottomrule
\end{tabular}
\end{table}

\end{document}